# The Harrington Yowlumne Narrative Corpus

Nathan M. White[*,1,2] and Timothy Henry-Rodriguez[1,2,3]

[1] Western Institute for Endangered Language Documentation
[2] Language and Culture Research Centre, James Cook University
[3] California State University, Fullerton

## Abstract

Minority languages continue to lack adequate resources for their development, especially in the technological domain. Likewise, the J.P. Harrington Papers collection at the Smithsonian Institution are difficult to access in practical terms for community members and researchers due to its handwritten and disorganized format. Our current work seeks to make a portion of this publicly-available yet problematic material practically accessible for natural language processing use. Here, we present the Harrington Yowlumne Narrative Corpus, a corpus of 20 narrative texts that derive from the Tejoneño Yowlumne community of the Tinliw rancheria in Kern County, California between 1910 and 1925. We digitally transcribe the texts and, through a Levenshtein distance-based algorithm and manual checking, we provide gold-standard aligned normalized and lemmatized text. We likewise provide POS tags for each lemmatized token via a lexicon-based deterministic approach. Altogether, the corpus contains 57,136 transcribed characters aligned with 10,719 gold standard text-normalized words.

## 1. Introduction

Minority languages continue to fall behind in the natural language processing and artificial intelligence boom of recent years, primarily due to a severe lack of available, appropriate data that could enable their development. Furthermore, many of these languages remain underdeveloped generally, and so often experience decline and extinction in favor of resource-rich languages, such as English. From a natural language processing point of view, this is a critical issue as modern state-of-the-art approaches to many tasks may ultimately be more a reflection of their superior performance just for English or some other major language than an actual breakthrough for language technology in general. Likewise, from a social standpoint, many minority language speakers may be unaware of the materials available for their language, due to issues of accessibility.

The current work seeks to contribute to the amelioration of these problems by presenting a corpus of narrative texts from an extremely low-resource minority language that today is considered by experts to be moribund or extinct, Yowlumne, but is subject to relatively significant revitalization efforts. This language belongs to the Yokuts group and is well-known among theoretical phonologists (as "Yawelmani") for its challenging morphophonological complexity, rendering it an important test language for any new phonological theory. Thus, Yowlumne represents an ideal language for testing new natural language processing approaches involving phonology such as an orthographic representation, especially as representative of issues common to low-resource languages, yet to date the language lacks any coherent corpus on which such research can be conducted.

However, raw data has existed for many years that could serve as the basis of such a corpus. J.P. Harrington conducted extensive fieldwork with the Tejoneño Yowlumne community at the Tinliw rancheria, focused primarily around collecting testimony and narrative evidence in favor of preserving their land rights in a court case in the early 1920s. The narrative texts contained in his collection are

---

invaluable representatives of texts in the Yowlumne language that, in their original form, are generally not accessible in practical terms to the community or to researchers. The corpus we present here are these texts, 20 in number, rendered in a more accessible and standardized format.

To the best of our awareness, the Harrington Yowlumne Narrative Corpus is the first gold-standard text normalization corpus for an extremely low-resource language, and especially for one with high morphophonological complexity.

## 2. Origin and Methodology

The Harrington Yowlumne Narrative Corpus is adapted from tarchives containing J.P. Harrington's collection as found on the Smithsonian Institution's website.[1] The pages that constitute his collection are in the form of scanned microfiche, uploaded slide by slide as separate viewable images. The locations in the microfiche slides of the texts that form part of the current corpus is provided in Appendix A. Each of these texts was digitally transcribed and annotated with normalized text with alignments to the original character sequences provided. The text normalization process took place over several stages. First, extraneous material found in the text, such as stray English glosses, were removed from the digital transcriptions. Second, a pre-annotation script utilizing Levenshtein distance (Levenshtein, 1966) provided the lexemes with the minimum distance based on the transcribed word form simplified to a series of symbol classes with the symbol classes of the transcribed word and that of each target word in a rudimentary dictionary constructed for this task. Some examples of the symbol classes appear in Table 1. The symbol series approach was adopted due to Harrington's tendency to transcribe the same phonemes in the same words in different ways depending on the distinctions he could hear and the degree of detail he chose to provide. The dictionary was based on lexemes found in Newman (1944), a wordlist created by William Shipley,[2] glosses that Harrington himself provide, and, in a handful of cases, a publicly-available community dictionary.[3]

The target words were limited to basic lexical forms for lexemes, while function words such as pronouns and demonstratives were permitted with all possible forms; this decision was made as some of these possible forms serve derivative functions as conjunctions and adverbs, and in some cases, these derivative functions are not distinguishable from their sources. Likewise, while the paradigms for pronouns and demonstratives have been fully documented by linguists, most lexemes lack full documentation of their paradigmatic forms, and thus the exact parsing of a form in many cases is subject to speculation that is best avoided.

The formalization of the algorithm used in the script is presented in Appendix B.

| class | examples of symbols |
|---|---|
| k | k k' g k` kh k̠ k̄ ḳ k' k: |
| r | tr tr' dr tr` tʃ ṭ t tʃ t' ṭ t' t̠r tʃr` tʃr |
| m | m m̄ ḿ m: |
| x | x q q̄ q̇ q́ |
| e | e é ē ɛ ɛ e: ḗ ɛ́ é: é ɛ́ ẹ é: ɛ: é: ɛ̣ ě e` |
| o | o o: ó ō ṓ ŏ ǫ o: ó ɔ ó: ɔɔ ɔ: ɔ̣ ɔ́ ʌ ʌ́ ǫ ǫ́ σ ò ȯ o` |
| y | y j í̠ i̠ y: j̠ |

Table 1. Examples of symbol classes.

The output of the automated pre-annotation script was hand-checked and extensively corrected by an expert linguist, with reference to Harrington's glosses and translations where available.

---

[1] https://collections.si.edu/search/detail/edanmdm:siris_arc_363758

[2] https://cla.berkeley.edu/item.php?bndlid=1696

[3] https://tulerivertribe-nsn.gov/wp-content/uploads/2019/01/ENGLISH-YOWLUMNI-DICTIONARY-Reduced.pdf

The performance of the Levenshtein distance-based pre-annotation script versus the gold-standard annotations appears in Table 2 below. Metrics provided include BLEU scores (Papineni et al., 2002) as well as the mean normalized Word Error Rate (WER; Klakow and Peters, 2002) approximated via the Normalized Edit Distance algorithm (Marzal and Vidal, 1993). These metrics have been chosen to best handle the spacing and word alignment issues found in the original texts as described in Section 4 below. The BLEU-1 metric in particular shows that the pre-annotations correctly identified tokens 68.04% of the time, while the WER indicates a mean error of 32.55%.

| Metric | Performance |
| --- | --- |
| WER | 0.3255 |
| BLEU-1 | 0.6804 |
| BLEU-2 | 0.5443 |
| BLEU-3 | 0.4422 |
| BLEU-4 | 0.3542 |

Table 2. Performance of Levenshtein distance-based pre-annotation.

Part-of-speech (POS) tags have been generated for each token in the corpus via the POS classification found in the rudimentary dictionary developed for the project based on looking up the POS tag by lemma. For those items not found in the dictionary or in the occasional case where the dictionary contained an inconsistent POS tag classification, the expert linguist specified the correct POS tag. The POS tags so generated were then checked for appropriate alignment with the entries.

The scheme for supplying POS tags relies on the following special features:

1) the tag UN is used for tokens classified as '[unknown]' as well as those tokens whose form was discernible by the expert linguist but for which the exact part of speech could not be determined due to lack of data;

| Tag | Class | Examples | # Tokens |
| --- | --- | --- | --- |
| CC | Conjunction | *yow* 'and, also, again', *wilshrin* 'but, in spite of that', *wa* 'however' | 297 |
| CD | Cardinal Number | *bonoy* 'two', *muno:shr* 'eight' | 13 |
| DT | Determiner | *traw* 'that.SG.LOC', *kin* 'this.SG.PRIM' | 1026 |
| FW | Foreign Word | *kale:ta'* 'cart', *mulo:* 'mule' | 78 |
| JJ | Adjective | *mani'* 'much, many', *c'olol* 'white' | 190 |
| NN | Common Noun | *yokoc'* 'person, people', *tinel* 'hole' | 1255 |
| NP | Proper Noun | *gaswaw* '[location]', *guchrchrun* '[name]' | 112 |
| ON | Onomatopoeia | *tw̥* '[sound]' | 1 |
| PN | Pronoun | *'amin* '3SG.GEN', *watuk* 'who.NOM' | 2010 |
| PP | Adposition-like Element (Weigel, 2005) | *p'e:p'at'iw* 'on top of', *'abiy* 'with.ACCOMP' | 50 |
| PU | Punctuation | . , | 1453 |
| RB | Adverb ("Particle" in Newman, 1944) | *hiya:m'i'* 'a long time ago', *canum* 'immediately' | 1417 |
| UH | Interjection | *họ:họ'* 'yes', *yah* 'hey!' | 21 |
| UN | Unknown | [unknown] | 335 |
| VB | Verb | *wiyi* 'say, do', *ta:na* 'go' | 2461 |
| | | **Total** | 10,719 |

Table 3. Distribution of POS tags in the corpus.

2) where words belong to more than one part-of-speech class due to derivational uses that historically stem from grammaticalization, as with *'ama'* '3SG.NOM; and' and *traw* 'that.SG.LOC; if', the originating part-of-speech class is used (e.g., PN for *'ama'* and DT for *traw*) to avoid potential ambiguity.
3) The POS tags mostly follow the Penn Treebank (Xue et al., 2005), with minor modifications as detailed below.

The distribution of the POS tags in the corpus are presented in Table 3.

## 3. Limitations of the Study

The creation of the annotated corpus faced three limitations. First, the derivational issues involving function words and the lack of documentation of paradigms of most lexemes limited the project as described in Section 2 above.

Second, as only one expert linguist was available for this project, given the extremely limited circumstances of this language at the present time, unfortunately no inter-annotator measures could be considered. In any case, however, a set of gold-standard alignments for text normalization was the result, in the sense that these have been hand-checked and confirmed by an expert in the language.

Third, the final corpus contains a special tag '[unknown]' in a number of locations where the original form represented by Harrington's transcription is not straightforwardly reconstructible, for one of three reasons:

1) Harrington's transcription is an impossible sequence in the language, suggesting a transcription error was made;
2) Harrington's transcription is ambiguous, potentially indicating one or more words; or
3) the word is not glossed or attested in any known source that provides a translation, meaning that it cannot be identified, since no fully fluent native speakers remain.

However, the vast majority of the forms found in the corpus were readily reconstructible and aligned with their text-normalized equivalents, producing a highly useful result. Likewise, where the type for each form could be identified, identification of the appropriate POS tag was straightforward.

## 4. Copyright and ethics considerations

The composition of the original handwritten documents took place more than 95 years ago, according to Harrington's notes and other documentation, which places them beyond the duration of copyright under U.S. law generally. The corpus that has resulted from the steps described in Section 2 reflects significant alteration of the original content in any case, such that one can deem a new work to have emerged to the extent that copyright is concerned.

With regard to ethics, J.P. Harrington is understood to have had the proper permissions to conduct the data collection in the first place as a government representative. No further work involving community participants took place to produce the corpus as presented here.

## 5. Corpus Format and Statistics

The corpus is made up of one file per narrative text, with each line generally corresponding to a gold-standard text-normalized word, with one word per line. Each word has five entries, separated by tabs:

1) the original character sequence as transcribed from Harrington's transcription;
2) the text-normalized form as described above;
3) the English gloss of the text-normalized form;
4) the linguistic expert's subjective certainty of the gloss provided (which is 100.0% in the vast majority of cases);

5) the POS tags for each text-normalized token.

Note that the following special cases are present. First, where the original character sequence contains spaces in inappropriate locations, an ampersand (&) appears at the end of the character sequence with the rest of the line blank to indicate that it must be joined with the character sequence in the next line to form the word. Second, where the original character sequence lacks spaces where a word boundary is present, a number sign (#) is placed at the word boundary with the other three tab-separated fields present, and the remaining portion of the character sequence moved to the next line with its own tab-separated fields as appropriate. An excerpt from the corpus illustrating this format appears in Figure 1 below.

| mi'n | mi'in | 'soon' | 100.0% |
| mak | mak' | '1DU.INCL.NOM' | 100.0% |
| ʃóqen | shroxo | 'exterminate, massacre' | 100.0% |
| jētr'aw | ye:tr'aw | 'all' | 100.0% |
| k`in | kin | 'this.SG.PRIM' | 100.0% |
| 'aʃ | 'ashr | 'actually, really' | 100.0% |
| tr`ḗ% | tri' | 'house' | 50.0% |
| phañi | pana: | 'arrive' | 50.0% |
| tr`aw | traw | 'that.SG.LOC' | 100.0% |
| mak | mak' | '1DU.INCL.NOM' | 100.0% |
| hi' | hi' | 'FUT' | 100.0% |
| wá% ts'u& | wa | 'however' | 50.0% |
| móno | c'o:mu | 'destroy, devour' | 50.0% |
| wijá'ăn | wiyi | 'say, do' | 100.0% |
| , | , | [punc] | 100.0% |

Figure 1. Corpus excerpt.

The general statistics for the corpus are found in Table 4.

| parameter | total count |
| --- | --- |
| total non-whitespace characters | 57,136 |
| mean characters / text | 2856.80 |
| total gold-standard words | 10,719 |
| mean words / text | 535.95 |
| total unique gold-standard words (incl. punctuation) | 792 |
| mean unique words / text | 39.6 |
| total sentences | 468 |
| mean sentences / text | 23.4 |

Table 4. General statistics.

## 6. Corpus Contributions

The Harrington Yowlumne Narrative Corpus contributes significantly to natural language processing efforts for low-resource languages in the following ways:

- it is the first publicly available text normalization dataaset for a highly-agglutinating low-resource minority language;
- it provides the first POS-tagged dataset for a California indigenous language;
- it provides the first text normalization dataset for a language without a normative orthographic standard, which represents the typical case for the vast majority of low-resource languages in need of normalization;
- it provides a dataset that truly represents a low-resource setting for research purposes, rather than a high-resource setting merely scaled down with ample additional resources upon which the task can rely;
- it provides a dataset for a language that is well known among theoretical phonologists as a pivotal language for testing phonological theories due to its morphophonological complexity, providing a challenge representative of the most intractable low-resource languages in terms of how words may be realized due to phonological issues;
- it provides a successful means of how to handle the tokenization

issues involving irregular spacing in non-normalized raw text for a low-resource language;
- it is publicly available to download; and
- it represents an effective, replicable paradigm for the development of additional corpora for other low-resource languages where noisy raw data is available.

## 7. Conclusion

We have created the first gold-standard text-normalization aligned corpus with POS tags for an extremely low-resource language, for a California indigenous language in particular, and derived from data stored in the J.P. Harrington Papers at the Smithsonian Institution. The approach to corpus construction described should prove highly replicable for other extremely low-resource languages for which archival materials are available, such as most indigenous languages of North America. The Harrington Yowlumne Narrative Corpus is publicly available to access and download online.[4]

## A. Source Locations of Texts

The 20 texts are found in the Harrington Smithsonian Papers microfiche scans in the following locations:

1. Reel 100, slides 1084–1087
2. Reel 100, slides 1101–1104
3. Reel 100, slides 1104–1108
4. Reel 100, slides 1109–1110
5. Reel 100, slides 1110–1117
6. Reel 100, slides 1121–1124
7. Reel 100, slides 1124–1126
8. Reel 100, slides 1128–1129
9. Reel 100, slides 1131–1134
10. Reel 100, slides 1134–1136
11. Reel 100, slides 1145–1154
12. Reel 101, slides 96–100
13. Reel 101, slides 101–105
14. Reel 101, slides 105–111
15. Reel 89, slides 500–502
16. Reel 89, slides 507–534
17. Reel 89, slides 534–542
18. Reel 89, slides 496–500
19. Reel 89, slides 503–507
20. Reel 100, slides 1119–1120

---

[4] http://corpus.ap-southeast-2.elasticbeanstalk.com/ywlcorpus/

## B. Text Normalization Algorithm

The formalization for the text normalization algorithm utilizing Levenshtein distance and symbol series appears in Algorithm 1 below.

1. Let
   $M$ represent the cardinality of tokens in the text,
   $D$ represent the dictionary,
   $S$ represent the matched output sequence, and
   $g$ represent the function calculating the Levenshtein distance
2. $S := \emptyset$
3. **for** $j \leftarrow 1$ to $m$ **do**
4.   **if** $w_j \in \{\text{punctuation}\}$, **then** $s_j := [\text{punc}]$
5.   **if** $w_j \notin \{\text{possible Yowlumne sound sequences}\}$, **then** $s_j := [\text{foreign}]$
6.   **else**
7.     $\tilde{w}_j \leftarrow \text{symbol\_series}(w_j)$
8.     **if** $\tilde{w}_j \in \{D\}$, **then** $s_j := d_{\tilde{w}_j}$
9.     **else**
10.       $\text{best} \leftarrow \underset{d}{\arg\max} \left( g(d_i, \tilde{w}_j) \right)$
    $\forall i \text{ s.t. } d_{i,0} = \tilde{w}_{j,0}$
11.       $\text{match\_score} \leftarrow \text{convert}(score_{best}, len(best))$
12.       **if** match_score $<= 0.7$, **then**
13.         stemmed $\leftarrow$ remove_final_suffix($\tilde{w}_j$)
14.         alternate $\leftarrow \underset{d}{\arg\max}(g(d_i, stemmed))$,
    $\forall i \text{ s.t. } d_{i,0} = stemmed_0$
15.         **if** $score_{alternate} < score_{best}$, **then** $s_j := $ alternate
16.         **else** $s_j := $ best
17. **end for**

Algorithm 1: Levenshtein with Symbol Series char2word normalization